%% file: paper.tex
\documentclass[acmtog,authorversion]{acmart} %,anonymous,review

\acmSubmissionID{805}

\usepackage{booktabs}
\usepackage{graphicx}
\usepackage{enumitem}
\usepackage{soul}
\usepackage{dsfont}
\usepackage{gensymb}
\usepackage{microtype}
\usepackage{pifont}

\usepackage{amssymb}
\usepackage{amsmath}
\usepackage{multirow}	
\usepackage{color, colortbl}
\usepackage{collect}
\usepackage{xspace}
\usepackage{makecell}

\input{definitions}

\copyrightyear{2024}
\acmYear{2024}
\setcopyright{rightsretained}
\acmConference[SA Conference Papers '24]{SIGGRAPH Asia 2024 Conference Papers}{December 3--6, 2024}{Tokyo, Japan}
\acmBooktitle{SIGGRAPH Asia 2024 Conference Papers (SA Conference Papers'24), December 3--6, 2024, Tokyo, Japan}\acmDOI{10.1145/3680528.3687655}
\acmISBN{979-8-4007-1131-2/24/12}

\begin{document}

\title{Manifold Sampling for Differentiable Uncertainty in Radiance Fields}

\author{Linjie Lyu}
\email{llyu@mpi-inf.mpg.de}
\orcid{0009-0007-4763-8457}
\affiliation{%
  \institution{Max-Planck-Institut für Informatik}
  \country{Germany}
}
\author{Ayush Tewari}
\email{ayusht@mit.edu}
\orcid{0000-0002-3805-4421}
\affiliation{%
  \institution{MIT CSAIL}
  \country{USA}
}
\author{Marc Habermann}
\email{mhaberma@mpi-inf.mpg.de}
\orcid{0000-0003-3899-7515}
\affiliation{%
  \institution{Max-Planck-Institut für Informatik}
  \country{Germany}
}
\author{Shunsuke Saito}
\email{shunsuke.saito16@gmail.com}
\orcid{0000-0003-2053-3472}
\affiliation{%
  \institution{Meta Codec Avatars Lab}
  \country{USA}
}
\author{Michael Zollhöfer}
\email{zollhoefer@meta.com}
\orcid{0000-0003-1219-0625}
\affiliation{%
  \institution{Meta Codec Avatars Lab}
  \country{USA}
}
\author{Thomas Leimkühler}
\email{thomas.leimkuehler@mpi-inf.mpg.de}
\orcid{0009-0006-7784-7957}
\affiliation{%
  \institution{Max-Planck-Institut für Informatik}
  \country{Germany}
}
\author{Christian Theobalt}
\email{theobalt@mpi-inf.mpg.de}
\orcid{0000-0001-6104-6625}
\affiliation{%
  \institution{Max-Planck-Institut für Informatik}
  \country{Germany}
}

\input{sec/0_abstract}

\begin{teaserfigure}
	\includegraphics[width=\textwidth]{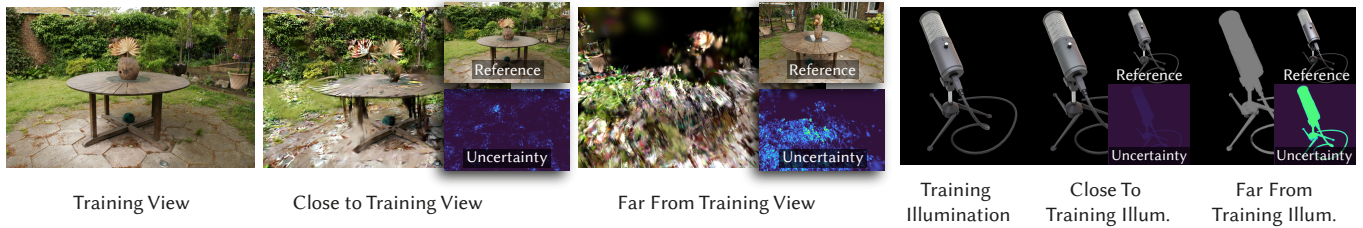}
	\caption{
        We propose a method to quantify uncertainty in radiance fields using a novel and efficient manifold sampling strategy. Our approach allows for differentiable optimization of views to reduce reconstruction ambiguities. We demonstrate this capability in next-best view planning (left) and illumination planning (right) tasks.
    }
	\label{fig:teaser}
	\Description[TeaserFigure]{TeaserFigure}
\end{teaserfigure}

\maketitle
\renewcommand{\shortauthors}{L. Lyu, A. Tewari, M. Habermann, S. Saito, M. Zollhöfer, T. Leimkühler, C. Theobalt}

\input{sec/1_intro}
\input{sec/2_relatedwork}

\input{sec/3_background}
\input{sec/4_method}
\input{sec/5_results}

\input{sec/6_conclusion}

%%%%%%%%%%%%%%%%%%%%%%%%%%%%%%%%%%%%%

\begin{acks}

This work was supported by the ERC Consolidator Grant 4DReply (770784).
\end{acks}

%%%%%%%%%%%%%%%%%%%%%%%%%%%%%%%%%%%%%

%{\small
% \clearpage
\bibliographystyle{ACM-Reference-Format}
\bibliography{paper}
%}
%\input{supplementary}
\end{document}

%% file: definitions.tex
\newcommand{\ie}{i.e.,\ }

\definecolor{aliceblue}{rgb}{0.94, 0.97, 1.0}
\newcolumntype{a}{>{\columncolor{aliceblue}}r}

\definecolor{junglegreen}{rgb}{0.113, 0.639, 0.5}

\newcommand{\rev}[1]{\textcolor{black}{#1}} 

\newcommand{\refFig}[1]{Fig.~\ref{fig:#1}}
\newcommand{\refTab}[1]{Tab.~\ref{tab:#1}}
\newcommand{\refSec}[1]{Sec.~\ref{sec:#1}}
\newcommand{\refEq}[1]{Eq.~\ref{eq:#1}}

\definecollection{mymaths}

\newcommand{\mymath}[2]{
    \newcommand{#1}{\TextOrMath{$#2$\xspace}{#2}}
    \begin{collect}{mymaths}{}{}{}{}
    #1
    \end{collect}
}

\mymath{\trainableparams}{{\boldsymbol{\theta}}}
\mymath{\radiancefield}{F_\trainableparams}
\mymath{\location}{\mathbf{x}}
\mymath{\direction}{\mathbf{d}}
\mymath{\auxcoords}{{\boldsymbol{\xi}}}
\mymath{\auxcoordsdim}{{n_\auxcoords}}
\mymath{\density}{\sigma}
\mymath{\rgb}{\mathbf{c}}
\mymath{\paramdim}{{d_{\trainableparams}}}
\mymath{\pixelcolor}{C}
\mymath{\ray}{\mathbf{r}}
\mymath{\rayorigin}{\mathbf{o}}
\mymath{\trainingdata}{\mathcal{T}}
\mymath{\numtrainingsamples}{{N_\trainingdata}}
\mymath{\image}{I}
\mymath{\uncertainty}{U}
\mymath{\volume}{V}
\mymath{\volumelossweight}{\lambda}
\mymath{\nummcsamples}{M}
\mymath{\generator}{G}
\mymath{\mean}{\boldsymbol{\bar{\theta}}}
\mymath{\genmatrix}{B}
\mymath{\covariance}{\Sigma}
\mymath{\volumesamples}{\mathbf{z}}
\mymath{\manifolddim}{k}
\mymath{\manifoldgenmatrix}{{\hat{\genmatrix}}}
\mymath{\manifoldvolumesamples}{{\hat{\volumesamples}}}
\mymath{\manifoldcovariance}{{\hat{\covariance}}}
\mymath{\transfermatrix}{T}

%% file: sec/0_abstract.tex
\begin{abstract}

Radiance fields are powerful and, hence, popular models for representing the appearance of complex scenes.
Yet, constructing them based on image observations gives rise to ambiguities and uncertainties.
We propose a versatile approach for learning Gaussian radiance fields with explicit and fine-grained uncertainty estimates that impose only little additional cost compared to uncertainty-agnostic training.
Our key observation is that uncertainties can be modeled as a low-dimensional manifold in the space of radiance field parameters that is highly amenable to Monte Carlo sampling.
Importantly, our uncertainties are differentiable and, thus, allow for gradient-based optimization of subsequent captures that optimally reduce ambiguities.
We demonstrate state-of-the-art performance on next-best-view planning tasks, including high-dimensional illumination planning for optimal radiance field relighting quality.

\end{abstract}

%% file: sec/1_intro.tex
\section{Introduction}
\label{sec:introduction}
% \AT{In the title, should we use radiance fields, given the relighting results, or should we make it more general?}\ShS{maybe neural fields?}
% \TL{don't people use the term "relightable radiance fields"?}
Recent years have witnessed the overwhelming success of volumetric radiance field representations for scene reconstruction and rendering~\cite{tewari2022advances,mildenhall2020nerf,kerbl3Dgaussians}.
Typically, such algorithms are provided with images observing the scene from multiple viewpoints, but also additional input variation such as time~\cite{park2021nerfies,xian2021space} or illumination~\cite{martin2021nerf,nerv2021} have been considered.

The reconstruction of 
%(possibly multi-dimensional) 
radiance fields 
%\ShS{is multi-dimensional referring to relighting case? calling it multi-dim RF sounds wrong to me.} 
from image observations is an instance of an ill-posed inverse problem:
Multiple different reconstructions can explain the data~\cite{tarantola2005inverse}.
Therefore, in addition to raw predictions, a complete and robust radiance field model must also provide a measure of its epistemic \emph{uncertainty}~\cite{pan2022activenerf,hoffman2023probnerf,shen2021stochastic,goli2023,Jiang2024FisherRF}. 
Such an uncertainty estimate should arguably exhibit three key properties:
First, it should provide a \emph{high-quality and expressive estimate} of the range of all scene attributes, avoiding proxy computations.
Second, the involved computations should be \emph{efficient} to not interfere with high-performance pipelines~\cite{kerbl3Dgaussians,fridovich2022plenoxels,muller2022instant}. %\AT{needs results to back it.}
Finally, the uncertainty estimate should be \emph{differentiable}. %\AT{Maybe move this point up?}
This crucial property turns uncertainty into a functional tool, as it allows to systematically increase model confidence by optimizing for viewpoints, lighting conditions, etc. for subsequent capture -- a particularly pressing challenge in resource-constrained applications.
Further, differentiability facilitates capture planning for high-dimensional input domains, where plain enumeration and evaluation of candidate capture conditions is infeasible.
In this work, we set out to develop a novel, general, and practical approach for uncertainty estimation in radiance fields that exhibits all the above desirable properties.

Existing methods for uncertainty quantification in radiance fields typically come in one of two flavors.
Stochastic approaches consider explicit attribute distributions which are optimized during training~\cite{shen2021stochastic,savant2024modeling,sunderhauf2023density,shen2022conditional,Yan2023iccv}, often within a variational-inference framework~\cite{blei2017variational}.
While they directly model distributions of parameters, such approaches often require many samples to obtain stable estimates, significantly impairing efficiency.
In a second line of work, uncertainty is estimated based on fully trained radiance fields~\cite{goli2023,Jiang2024FisherRF}. %\AT{Isn't \cite{pan2022activenerf} variational inference?}.
Typically, a Laplace approximation~\cite{daxberger2021laplace,ritter2018scalable} is employed, resulting in rather coarse uncertainty proxies.
Since these estimates require automatic differentiation, obtaining differentiable uncertainties necessitates the computation of higher-order derivatives, posing significant practical challenges.

Our approach employs a stochastic radiance field~\cite{shen2021stochastic,savant2024modeling} with a 3D Gaussian representation~\cite{kerbl3Dgaussians} that treats all of its parameters as random variables.
Drawing samples from the joint distribution of parameters allows for rendering different realizations of the radiance field.
Both, training and uncertainty estimation during inference, %\LL{do we want to make it clear it's for inference stage?}, 
require an integral over realizations, which we estimate using Monte Carlo sampling.
However, the high dimensionality of radiance field parameters poses a significant challenge, as the joint distribution is typically complex.
Incorporating complete covariance matrices across all parameters is intractable.
Assuming full independence between parameters ~\cite{shen2021stochastic,pan2022activenerf,savant2024modeling}, even though common practice for efficiency, %\mhc{whats meant here by common practice? Other approaches typically assume it even though its wrong?},
leads to excessive variance.
Our key contribution is the observation that restricting samples to a \emph{low-dimensional linear manifold in parameter space} is sufficient for an Monte Carlo estimator to efficiently train and estimate uncertainty in a stochastic radiance field.
While stochastic sampling is typically considered inefficient and prone to high variance, with our method drawing only very few samples is enough to obtain stable and high-quality gradients for training and uncertainty estimation (\refFig{teaser}).

The benefits of our formulation are threefold.
First, explicit modeling of the parameter distribution provides a convenient and interpretable uncertainty measure.
Second, the \emph{low-rank approximation} of the
covariance matrix results in a smoother objective energy landscape that lends itself to stable optimization. 
% \mhc{"stable to dune" -> unclear. also why "implicitly"? maybe say: Thus, the energy landscape becomes implicitly smoother ensuring better convergence during optimization.} 
Further, due to the low number of samples required, our strategy is highly efficient in terms of computational performance and memory requirements, both during training and inference. 
Third, our uncertainties are trivially differentiable, as they simply arise from the mean
% \mhc{MC estimate rather than sum?} 
of rendered radiance field realizations.

We demonstrate that our approach significantly outperforms previous methods across multiple aspects.
A main application of our manifold sampling is next-best view planning, which we evaluate on a variety of relevant scenarios.
In addition to the optimal selection of camera candidates~\cite{pan2022activenerf,kopanas2023,Jiang2024FisherRF}, an application where we outperform the state of the art, our approach \emph{for the first time} allows for a \emph{differentiable} fine-grained optimization of camera parameters that optimally reduce uncertainty given the current state of the model.
We further show that our approach enables differentiable uncertainty estimation in augmented radiance fields, exemplified by the task of relightable reconstruction~\cite{martin2021nerf,nerv2021}.
Concretely, we optimize for the next-best illumination condition that minimizes relighting uncertainty in radiance fields, demonstrating the versatility of our approach and its ability to handle high-dimensional domains.

In summary, our contributions are:

\begin{itemize}
    \item A novel stochastic radiance field formulation based on manifold sampling that can be efficiently trained.
    \item A method for differentiable uncertainty estimation that allows for fine-grained optimization of subsequent scene captures.
    \item The evaluation of our approach on the tasks of next-best viewpoint and illumination planning.
\end{itemize}

\noindent
% We will make all source code available on publication.
 We provide all source code and pre-trained models on \url{https://vcai.mpi-inf.mpg.de/projects/2024-ManifoldUncertainty/}.

%% file: sec/2_relatedwork.tex
\section{Related Work}
\label{sec:relatedwork}
%\AT{Please add names of any relevant papers that I missed.}
\subsection{Deterministic 3D Reconstruction} %\ShS{perhaps, we should add references?}
3D reconstruction is the problem of taking image observations as input and reconstructing the underlying 3D scene.
%that can be used to generate images from novel scene conditions, such as novel camera viewpoints or lighting conditions. 
%
This problem has been studied for a long time and mature systems such as structure-for-motion (SfM) are widely used to reconstruct 3D geometry and cameras from image observations~\cite{colmap}. 
Recent years have seen a lot of progress on 3D scene representations that enable high-quality novel view synthesis. 
Early advances relied on neural scene representations, such as neural radiance fields~\cite{mildenhall2020nerf}, which use neural networks to represent geometry and appearance quantities, and are optimized from the input images using inverse volume rendering. 
More recently, voxel-based representations~\cite{yu2021plenoctrees,fridovich2022plenoxels,kplanes_2023,muller2022instant} and 3D Gaussians~\cite{kerbl3Dgaussians} have been used to represent the underlying scene for efficient reconstruction and rendering. 
Similar progress has also been seen for relighting problems, where differentiable physics-based rendering, or image-based lighting formulations have been used to compute 3D reconstructions~\cite{zhang2021nerfactor,chen2024urhand,saito2024rgca,zhang2021neural,mai2023neural,lyu2022nrtf}. 

Most of the progress has focused on \emph{deterministic} 3D reconstruction, where the underlying epistemic uncertainties are not taken into account. 
Instead, these methods rely on a large number of input observations to minimize the uncertainty. 
In contrast, we are interested in modeling and reconstructing the uncertainties in the problem. 
%
% original 3DGS \cite{kerbl3Dgaussians}

% original NeRF \cite{mildenhall2020nerf}

% local light field fusion \cite{mildenhall2019llff}

\subsection{Uncertainty in 3D Reconstruction}
%
% \LL{Should we narrow down the subsection title a little bit? E.g. Uncertainty in radiance field. 'Uncertainty' sounds too wide.  }
Modeling uncertainty is an active area of research in machine learning. 
%
% Common approaches are ensemble-based and Bayesian learning-based. 
%
% We focus our discussion on Bayesian learning-based methods. 
%
A common way of estimating uncertainty in deep learning is using variational inference, where a distribution over free parameters is estimated
% \LL{ 1.“Weight
% Uncertainty in Neural Networks” ; 2. Fast and Scalable Bayesian Deep Learning by WeightPerturbation in Adam; 3.Noisy Natural Gradient as
% Variational Inference}
~\cite{kendall2017uncertainties,khan2018fast,blundell2015weight,zhang2018noisy}. 
However, variational inference is often very slow and expensive, e.g., representing the full covariance matrix can be infeasible for larger models. 
Thus, many methods use Laplace's approximation 
% \LL{1.Transforming neural-net output
% levels to probability distributions; 2. Bayesian model comparison and backprop nets } 
that estimates the posterior distribution from the optimized MAP solution
% \LL{“1.A scalable laplace approximation for neural
% networks”;  Being Bayesian, Even Just a Bit,
% Fixes Overconfidence in ReLU Network }
~\cite{foong2019between,denker1990transforming,mackay1991bayesian,ritter2018scalable,kristiadi2020being,savant2024modeling}.
While Laplace's approximation is more efficient, it can struggle with recovering complex distributions. 
% \ShS{maybe we should explain why Laplace's approximation is bad. this is not clear in this paragraph}
%

Both strategies have also been explored in the context of radiance fields and 3D reconstruction. 
Variational inference-based approaches compute the posterior distribution.
S-NeRF~\cite{shen2021stochastic}, ActiveNeRF~\cite{pan2022activenerf}, and \citet{savant2024modeling} assume independence between all parameters of the 3D representation.  
% \LL{1.modeling uncertainty for Gaussian Splatting (they model per gaussian local covariance)}
%
CF-NeRF~\cite{shen2022conditional} uses latent space modeling with normalizing flow models to learn more complex distributions. 
FisherRF~\cite{Jiang2024FisherRF} and Bayes' Rays~\cite{goli2023} use Laplace's approximation to compute the uncertainty in reconstruction. 
% FisherRF, Bayes' rays, 
Our method is based on variational inference; however, we show that a simple low-dimensional approximation of the covariance leads to tractable training and inference. 
Our method outperforms methods that assume independence between all parameters, and at a comparable cost to methods that use Laplace's approximation.
% \ShS{can we more explicitly explain the difference? like performs comparably with full rank covariance while being as fast as Laplace's approximation.}

We contrast our approach with popular 3D reconstruction methods that rely on dense training data~\cite{hoffman2023probnerf,tewari2024diffusion,long2024wonder3d,xu2023dmv3d,zeronvs}. 
These approaches assume large and dense training datasets with no uncertainty, which can then be used to recover epistemic uncertainty at test time from sparse observations. 
Our method does not rely on any dense training data, and only uses the sparse observations to compute the reconstruction. 

\subsection{Active 3D Reconstruction}
Reconstructing estimates of the uncertainty is important for active scanning applications, where the uncertainty can guide the optimal scene conditions, such as camera poses and lighting. 
The most common application is camera selection, where the sequence of cameras used for scanning is determined by the current estimate of the uncertainty~\cite{Jiang2024FisherRF,pan2022activenerf,kopanas2023}.
We compare to the state-of-the art methods and demonstrate better performance on this task. 
Further, we also introduce a differentiable version of the task where cameras are allowed to freely move in 3D, rather than restricted to some candidate cameras. 
We also demonstrate results on lighting optimization where we reconstruct scene appearance from very few lighting conditions. 
To the best of our knowledge, we are the first to demonstrate this application.
% \AT{All active slam + heuristics-based(kopanas) + fisherrf + (any for lighting)?}

% Bayes rays \cite{goli2023}: overlay a grid-based deformation field, determine the deformation using differential quantities, Laplace, etc.

% FisherRF \cite{Jiang2024FisherRF}: estimate the Fisher information (Hessian)

% ActiveNeRF \cite{pan2022activenerf}

% ProbNerf \cite{hoffman2023probnerf}

% Activermap \cite{zhan2022activermap}

% Ensembling \cite{sunderhauf2023density}

% Finding Waldo \cite{skartados2024finding}

% Active neural mapping \cite{Yan2023iccv}: MC estimation of uncertainty

% CG-SLAM \cite{hu2024cg}

% Stochastic neural radiance fields \cite{shen2021stochastic}: uncorrelated MC + VI (nerf)
% Modeling uncertainty for Gaussian splatting \cite{savant2024modeling}: uncorrelated MC + VI (3DGS)

% Conditional-flow NeRF \cite{shen2022conditional}

% Yorgos heuristics \cite{kopanas2023}

% low-rank covariance might be an instance of low-rank sampling / rank-1 lattices \cite{niederreiter1992random}

% sampling on a manifold in different contexts: \cite{Jakob2012Manifold}

%% file: sec/3_background.tex
\section{Background}
\label{sec:background}

In this work, we are concerned with radiance fields \radiancefield that model a 3D scene using the volumetric representation
\begin{equation}
\label{eq:radiance_field_definition}
    \radiancefield:
    (\location, \direction, \auxcoords)
    \rightarrow
    (\density, \rgb).
\end{equation}
The radiance field outputs volumetric density $\density \in \mathds{R}_+$ and RGB color $\rgb \in \mathds{R}^3$ as a function of three arguments:
$\location \in \mathds{R}^3$ is a location in the scene,
$\direction \in \mathds{S}^2$ is a 3D direction vector,
and $\auxcoords \in \mathds{R}^\auxcoordsdim$ is an (optional) vector of auxiliary parameters that depend on the application.
For example, \auxcoords is a scalar time dimension in the case of dynamic reconstruction~\cite{park2021nerfies,xian2021space}, or a high-dimensional vector of illumination conditions in the case of relightable scenes~\cite{martin2021nerf,nerv2021,lyu2022nrtf}.

How to best represent \radiancefield remains a subject of active research~\cite{mildenhall2020nerf,fridovich2022plenoxels,kerbl3Dgaussians,xu2022point,muller2022instant,yu2021plenoctrees,chan2022efficient}.
In this work, we focus on the 3D Gaussian Splatting (3DGS) representation~\cite{kerbl3Dgaussians}, which marks the current state of the art in terms of reconstruction quality and rendering speed.
This model represents a radiance field as a Gaussian mixture model, where each primitive has a location in 3D space, an anisotropic covariance, an opacity, and a view-dependent color represented in the spherical harmonics (SH) basis.
The detailed workings of this model are not central to our approach.
Therefore, we treat \radiancefield as a black box in our exposition and denote all trainable parameters as $\trainableparams \in \mathds{R}^{\paramdim}$.
Typically, a high number of parameters is required to parameterize a radiance field, with \paramdim often reaching into the millions.

A radiance field can be rendered to an RGB image $\image(\auxcoords)$ using emission--absorption volume rendering~\cite{kajiya1984ray}:
\begin{equation}
\label{eq:volume_rendering}
    \pixelcolor_\trainableparams(\ray, \auxcoords)
    =
    \int_{t_n}^{t_f}
    \exp \left(
        -
        \int_{t_n}^t
        \density \left( \ray(s), \auxcoords \right)
        \mathrm{d} s
    \right)
    \density \left( \ray(t), \auxcoords \right)
    \rgb \left( \ray(t), \direction, \auxcoords \right)
    \mathrm{d}t.
\end{equation}
Here, $\pixelcolor_\trainableparams$ is the color of a rendered pixel of \image corresponding to a camera ray
$\ray(t) = \rayorigin + t \direction$
with near and far bounds $t_n$ and $t_f$, respectively.
Again, the subscript \trainableparams denotes its dependence on the trainable parameters of \radiancefield.
The primitive-based 3DGS representation solves this integral using an approximation based on rasterization~\cite{zwicker2001ewa}.
Importantly, the rendering process is inherently differentiable, which enables the optimization of radiance field parameters \trainableparams such that renderings match a set of posed training views.

\begin{figure}[t]
    \includegraphics[width=\linewidth]{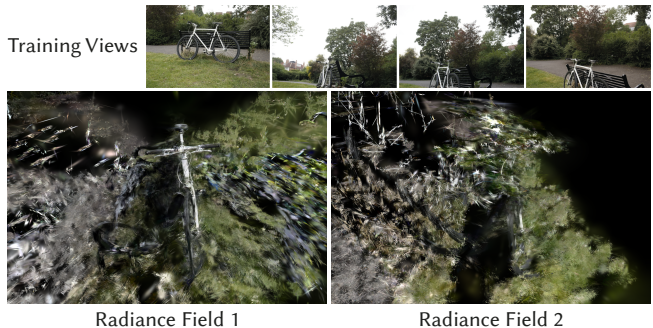}
    \vspace{-5mm}
    \caption{
    Using the same four training views (top), different model parameter initializations yield different radiance fields (bottom), reflecting uncertainty.
    }
    \label{fig:uncertainty_illustration}
\end{figure}

However, regrettably, obtaining a potentially high-dimensional radiance field from image observations alone is an ill-posed problem with nuanced ambiguities:
A whole distribution of radiance fields can result in the same rendered image.
This is illustrated in \refFig{uncertainty_illustration}, where two static radiance fields are trained using the same four training views (\refFig{uncertainty_illustration}, top) but starting from different initializations. 
Despite both radiance fields nearly perfectly fitting the training views, rendering the reconstructions from a test view reveals vastly different solutions (\refFig{uncertainty_illustration}, bottom).
This problem is even more pronounced in higher-dimensional settings, such as relightable radiance fields, where capturing multiple illumination conditions alongside multiple camera poses is necessary.
The degree of ambiguity gradually diminishes as more training views are captured.
However, finding an \emph{optimal} capture sequence requires accounting for scene-specific uncertainties.
In the following section, we develop an efficient approach to quantify these uncertainties, enabling us to plan the next-best observations that optimally increase model confidence.

%% file: sec/4_method.tex
\section{Method}
\label{sec:method}

The central goal of this work is to learn a (multi-dimensional) radiance field \radiancefield in the form of \refEq{radiance_field_definition} from image data that provides a fine-grained estimate of its inherent epistemic uncertainty.
We are especially focused on computing uncertainty estimates that are both efficient and differentiable, enabling their application in multi-dimensional next-best-view planning scenarios.
To this end, we devise a stochastic radiance field formulation (\refSec{stochastic_radiance_field}) with a low-dimensional manifold sampler at its core (\refSec{manifold_sampling}) that allows highly efficient training (\refSec{training}) and differentiable uncertainty estimation (\refSec{differentiable_uncertainty_optimization}).

We assume we have access to training data in the form of one or multiple posed images, providing per-pixel triplets
$
\trainingdata
=
\left\{
\ray_i, \auxcoords_i, \pixelcolor_i
\right\}_{i=1}^\numtrainingsamples
$.
Note how this setting is a two-fold generalization of typical radiance field reconstruction tasks.
First, the additional auxiliary parameter vector \auxcoords accommodates arbitrary and potentially high-dimensional scene configurations, such as different lighting conditions.
Second, we do not necessarily require multiple training images. 
Our algorithm is capable of handling a \emph{single} image as the initial input and utilizes uncertainty estimates to suggest which images to capture next for optimal successive uncertainty reduction.

%---------------------------------------------------------------

\subsection{Stochastic Radiance Field}
\label{sec:stochastic_radiance_field}

We consider a distribution of radiance fields $p(\radiancefield)$ which we realize using a probabilistic model, \ie by explicitly modeling the joint distribution $p(\trainableparams)$ of all trainable parameters.
Drawing a sample 
$\trainableparams^* \sim p(\trainableparams)$
gives rise to a specific radiance field realization $\radiancefield^*$.
\rev{Our objective is to optimize the posterior distribution $p(\radiancefield | \trainingdata)$ by minimizing rendering errors for training views while concurrently maximizing uncertainty for novel views. This approach aims to} obtain a posterior distribution that is as spread out as possible while making sure that \emph{each} realization $\radiancefield^*$ explains the training data \trainingdata, \ie when $\radiancefield^*$ is rendered using inputs $\ray_i$ and $\auxcoords_i$ from the training data set, it consistently outputs the corresponding pixel color $\pixelcolor_i$.

Since the dimensionality \paramdim is very high -- radiance fields routinely involve millions of parameters --, it is essential to take special care to maintain tractability.
We employ a variational-inference framework~\cite{blei2017variational} and choose to approximate $p(\trainableparams)$ using a continuous \emph{uniform} distribution.
In essence, this approximation models a compact hyper-volume $\volume \subset \mathds{R}^\paramdim$ within parameter space with a constant probability density:
\begin{equation}
    p(\trainableparams)
    =
    \begin{cases}
        \frac{1}{|V|} & \trainableparams \in \volume, \\
        0 & \textrm{else}.
    \end{cases}
\end{equation}
As the ground-truth distribution of parameters is generally unknown, we argue that choosing a uniform prior is a reasonable choice.
Although a Gaussian prior is commonly favored due to its mathematical convenience~\cite{pan2022activenerf}, %we do not find unimodality assumptions to be essential.
\rev{we do not use a Gaussian prior because it assumes a higher probability around the mean. A counter-example is a single-view case, where Gaussians can move along the depth. Their probability distribution along depth should not be symmetric and centered around any particular depth value.  Also, the color of occluded Gaussians should be equally likely to be of any value.
In addition, a uniform prior avoids sampling unbounded values such as Gaussian scales.
}
Central to our approach is a novel method for constructing the hyper-volume \volume, as described in \refSec{manifold_sampling}.

Using a uniform approximation of $p(\trainableparams)$, our training loss for finding an optimal \volume can be formulated as follows:
\begin{equation}
\label{eq:continuous_training_loss}
    \mathcal{L}
    =
    \sum_{i=1}^\numtrainingsamples
    \int_{\trainableparams^* \in \volume}
    \left\|
        \pixelcolor_{\trainableparams^*}(\ray_i, \auxcoords_i) - \pixelcolor_i
    \right\|_\image
    \mathrm{d} \trainableparams^*
    -
    \volumelossweight |\volume |.
\end{equation}
Here, the first term encourages that all realizations of \radiancefield match the training data, while the second term encourages a large hyper-volume.
The scalar \volumelossweight balances the two terms.

Given a trained radiance field, our main application of next-best-view planning requires an uncertainty estimate \uncertainty per image $\image(\auxcoords)$.
Similar to our training objective in \refEq{continuous_training_loss}, this can be formalized as an integral over rendered realizations of \radiancefield:
\begin{equation}
\label{eq:continuous_uncertainty}
    \uncertainty(\image(\auxcoords))
    =
    \sum_{\ray \in \image}
    \int_{\trainableparams^* \in \volume}
    \left\|
    \pixelcolor_{\trainableparams^*}(\ray, \auxcoords)
    -
    \bar{\pixelcolor}(\ray, \auxcoords)
    \right\|^2
    \mathrm{d} \trainableparams^*,
\end{equation}
where 
\begin{equation}
\label{eq:continuous_mean}
    \bar{\pixelcolor}(\ray, \auxcoords)    
    =
    \int_{\trainableparams^* \in \volume}
    \pixelcolor_{\trainableparams^*}(\ray, \auxcoords)
    \mathrm{d} \trainableparams^*
\end{equation}
is the mean pixel color across realizations.
A high uncertainty \uncertainty for a particular $\image(\auxcoords)$ indicates a good candidate for subsequent capture.

Clearly, the high-dimensional integrals in Eqs.~\ref{eq:continuous_training_loss}-\ref{eq:continuous_mean} over radiance field realizations cannot be evaluated analytically due to the nonlinearity of the rendering function.
Instead, we opt for a numerical integration scheme.
The most general approach that, in principle, scales to high dimensions is Monte Carlo integration, which employs a random sampling of the integration domain.
A Monte Carlo estimator of \refEq{continuous_training_loss} is given by
\begin{equation}
\label{eq:mc_training_loss}
    \mathcal{L}_\textrm{MC}
    =
    \sum_{i=1}^\numtrainingsamples
    \frac{1}{\nummcsamples}
    \sum_{\trainableparams^* \sim \volume}
    \left\|
        \pixelcolor_{\trainableparams^*}(\ray_i, \auxcoords_i) - \pixelcolor_i
    \right\|
    -
    \volumelossweight |\volume |,
\end{equation}
where $\trainableparams^*$ are now \nummcsamples random samples.
Note how \refEq{mc_training_loss} optimizes over the space of generative models that allow to draw samples from and thereby implicitly define \volume.
An analogous estimator for \refEq{continuous_uncertainty} is
\begin{equation}
\label{eq:mc_uncertainty}
    \uncertainty_\textrm{MC}(\image(\auxcoords))
    =
    \sum_{\ray \in \image}
    \frac{1}{\nummcsamples}
    \sum_{\trainableparams^* \sim \volume}
    \left\|
    \pixelcolor_{\trainableparams^*}(\ray, \auxcoords)
    -
    \bar{\pixelcolor}(\ray, \auxcoords)
    \right\|^2.
\end{equation}
The integral in \refEq{continuous_mean} can be estimated correspondingly.

The benefit of learning \radiancefield per \refEq{mc_training_loss} and estimating its uncertainty per \refEq{mc_uncertainty} is that is only requires a sum of first-order differentiable rendered radiance fields, \ie different from other approaches~\cite{goli2023, Jiang2024FisherRF}, our estimates are derivative-free.
This is particularly beneficial in the uncertainty estimate of \refEq{mc_uncertainty}, as it easily allows to differentiate $\uncertainty_\textrm{MC}$ with respect to camera parameters and/or \auxcoords to perform gradient-based optimization of next-best views, as described in \refSec{differentiable_uncertainty_optimization}.
However, regrettably, Monte Carlo estimators typically exhibit excessive variance and require a high number \nummcsamples of samples to provide stable results, in particular in high dimensions.
The key technical contribution of this work is the observation that the probability volume \volume can be modeled as a low-dimensional manifold.
This allows stable and high-quality training per \refEq{mc_training_loss} and uncertainty estimation per \refEq{mc_uncertainty} using a very low number of samples, as described next.

%---------------------------------------------------------------

\subsection{Manifold Sampling}
\label{sec:manifold_sampling}

We seek to devise a generative model that allows us to sample from a compact radiance-field-specific volume $\volume \subset \mathds{R}^\paramdim$ of constant probability density that captures the distribution of trainable parameters \trainableparams.
Ideally, we would aim for maximum expressivity by allowing \volume to have arbitrary shapes.
Powerful generative models of radiance field parameters have been studied~\cite{chan2021pi,gu2022stylenerf,muller2023diffrf} and typically rely on deep neural networks.
Unfortunately, our efficiency constraints -- training a radiance field needs to be quick while training a deep generative model on top is slow -- and the extremely high dimensionality \paramdim of the parameter space, make such solutions impractical. 
Therefore, we impose a more rigid structure.

\begin{figure}[h]
    \includegraphics[width=0.99\linewidth]{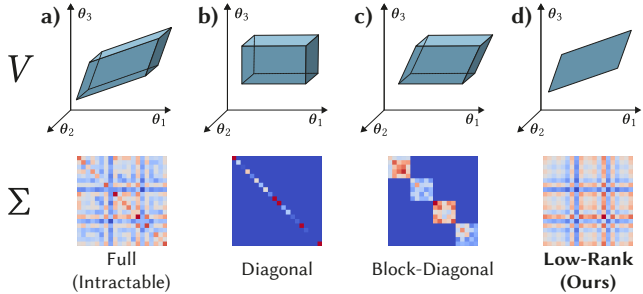}
    \vspace{-2mm}
    \caption{Different variants to model an uncertainty volume \volume in the space of radiance field parameters (top row, only three out of millions of parameters are shown) using different covariance matrices \covariance (bottom row, 20 dimensions are shown).
    \emph{(a)} A full \covariance is the most expressive solution that leads to an arbitrarily shaped parallelotope, but it suffers from an intractable number of parameters.
    \emph{(b)} Restricting \covariance to a diagonal matrix is a sparse solution, but it can only represent axis-aligned hyper-rectangles.
    \emph{(c)} A block-diagonal \covariance is slightly more expressive, but it requires making representation-specific independence assumptions and small blocks to stay tractable.
    \emph{(d)} Our solution employs a low-rank covariance matrix, which results in a manifold parallelotope (here a 2D parallelogram). This parameterization is highly efficient to train and results in expressive uncertainty estimates.}
    \label{fig:sampling_strategies}
\end{figure}

Towards our solution, we consider a linear model \generator of the form
\begin{equation}
\label{eq:linear_model}
    \generator(\volumesamples)
    =
    \mean +  \genmatrix \volumesamples,
\end{equation}
where $\mean \in \mathds{R}^\paramdim$ is the mean of the distribution, and 
$\volumesamples \sim \mathcal{U}( \left[-1, 1 \right]^\paramdim )$,
is uniform-randomly sampled.
The generating matrix 
$ \genmatrix \in \mathds{R}^{\paramdim \times \paramdim}$ 
encodes a linear relationship between parameter dimensions and defines the shape of \volume as a parallelotope (\refFig{sampling_strategies}a, top).
The covariance matrix of the emerging distribution of parameter samples is given by
$\covariance =  \genmatrix  \genmatrix^T \in \mathds{R}^{\paramdim \times \paramdim}$
(\refFig{sampling_strategies}a, bottom).
Trivially, a uniform sampling of \volumesamples from the hypercube results in a uniform sampling of \volume.
The most challenging aspect of \refEq{linear_model} lies in representing the generating matrix  \genmatrix, which, in general, involves a number of entries that grows with the \emph{square} of the number of trainable parameters -- a completely intractable quantity.

A commonly used solution to avoid this quadratic complexity is to presume \emph{independence} among the components of \trainableparams~\cite{shen2021stochastic,pan2022activenerf,goli2023, Jiang2024FisherRF}. 
In this case,  \genmatrix and, consequently, \covariance, reduce to a diagonal matrix, and \volume transforms into an axis-aligned hyper-rectangle (\refFig{sampling_strategies}b).
Although this strategy notably diminishes complexity, we observe a disproportionate compromise in expressivity and, thus, accuracy.
A different, slightly less invasive strategy would be to presume independence between \emph{groups} of variables.
In such a model,  \genmatrix and, consequently, \covariance, become block-diagonal and parameterize a parallelotope with a restricted distribution of admissible slopes (\refFig{sampling_strategies}c).
While this solution is strictly more expressive than using a diagonal matrix, we observe that it still suffers from reduced predictive accuracy.
Further, it requires making representation-specific independence assumptions between small groups of variables to retain tractability.

Our key observation is that for high-quality and efficient uncertainty quantification in radiance fields, \volume can be modeled as a linear \manifolddim-dimensional \emph{manifold} in \paramdim-dimensional parameter space, where
$\manifolddim \ll \paramdim$.
Intuitively, the uncertainty volume of radiance fields exhibits significantly fewer degrees of freedom compared to the number of parameters defining the radiance field itself.
Concretely, the variation of radiance field parameters \trainableparams can be well explained by a \manifolddim-dimensional parallelotope (\refFig{sampling_strategies}d, top).
This translates into a model
\begin{equation}
\label{eq:manifold_model}
    \generator(\manifoldvolumesamples)
    =
    \mean + \manifoldgenmatrix \manifoldvolumesamples,
\end{equation}
where
$ \manifoldgenmatrix \in \mathds{R}^{\paramdim \times \manifolddim}$
is the generating matrix of the manifold parallelotope, and 
$\manifoldvolumesamples \in \mathds{R}^\manifolddim$
are low-dimensional random samples.
Note how this construction results in a low-rank covariance matrix
$\manifoldcovariance =  \manifoldgenmatrix  \manifoldgenmatrix^T$ (\refFig{sampling_strategies}d, bottom).
Surprisingly, we observe that setting $\manifolddim = 2$ is sufficient for high-quality training and uncertainty estimation in practice, while also enabling fast training.

The advantages of this formulation are three-fold:
First, the low-rank approximation results in a compact representation, as we only require 
$(\manifolddim + 1) \times \paramdim$ parameters to represent our full model.
Second, \refEq{manifold_model} is very efficient to compute.
Third, the low effective dimensionality of \volume makes the Monte Carlo estimators in \refEq{mc_training_loss} and \refEq{mc_uncertainty} highly effective and practically noise-free.
We provide details on training and uncertainty estimation of our approach in \refSec{training} and \refSec{differentiable_uncertainty_optimization}, respectively.

%---------------------------------------------------------------

\subsection{Training}
\label{sec:training}

To train our stochastic radiance field, we incorporate \refEq{manifold_model} into \refEq{mc_training_loss} to arrive at our final training objective, which optimizes over the parameters \mean and \manifoldgenmatrix of our generative model:
\begin{equation}
\label{eq:our_training_objective}
    \mathcal{L}_\textrm{manif}
    =
    \sum_{i=1}^\numtrainingsamples
    \frac{1}{\nummcsamples}
    \sum_{\manifoldvolumesamples \sim \mathcal{U}( \left[-1, 1 \right]^\manifolddim )}
    \left\|
        \pixelcolor_{\generator(\manifoldvolumesamples)}(\ray_i, \auxcoords_i) - \pixelcolor_i
    \right\|
    -
    \volumelossweight 
    \| \manifoldgenmatrix \|_1.
\end{equation}
In the data term, we use the generator \generator to sample from the parameter manifold.
In practice, we use a low-discrepancy \citet{sobol1967distribution} sequence to sample \manifoldvolumesamples for improved stability.
\rev{We follow~\citet{kerbl3Dgaussians} and measure pixel differences using a sum of $\ell_1$ distance and the SSIM~\cite{wang2004image} metric, while also employing their heuristic densification strategies.}

The second term in \refEq{our_training_objective} is an estimator of the volume of \volume which we seek to maximize.
While the volume of an ordinary parallelotope is easily computed using the determinant of its generating matrix, the volume of a manifold of the shape of a parallelotope in a higher-dimensional ambient space is not straightforward.
As a substitute, we employ the entrywise L1 norm of \manifoldgenmatrix, \rev{promoting longer sides of the parallelotope,} which we found provides higher-quality results than all alternatives we tested. \rev{We choose the L1 norm over L2 because it provides a constant gradient, preventing both gradient vanishing when $\| \manifoldgenmatrix \|_1$ is small and gradient explosion when $\| \manifoldgenmatrix \|_1$ is large.
}

We observe that, for optimal quality, special attention must be given to the parameterization of \manifoldgenmatrix.
A na\"ive parameterization, \ie an unconstrained optimization of the entries of \manifoldgenmatrix, is prone to producing degenerate results where the columns of \manifoldgenmatrix are nearly linearly dependent.
As a result, the manifold volume can have a lower effective dimensionality than \manifolddim.
In our solution, we prevent such outcomes using a simple strategy:
The individual entries of \manifoldgenmatrix are initialized to low positive values, with a subsequent application of a ReLU activation function.
The result is multiplied with a random but fixed sign, \ie $+1$ or $-1$, per matrix entry.
This strategy effectively diversifies the directions of the column vectors of \manifoldgenmatrix, preventing degenerate solutions in practice.

Thanks to our manifold sampling, we find that a \emph{single} Monte Carlo sample per training iteration, \ie $\nummcsamples=1$, is sufficient for fast and stable convergence.
Notably, we observe that our approach does not increase the number of training iterations required to reach a converged result.
Consequently, \emph{our formulation imposes only minor extra cost compared to standard radiance field training}. 

Except for our loss in \refEq{our_training_objective} and a $(\manifolddim+1)$-fold increase in the number of trainable parameters, the training process follows the standard approach described in \citet{kerbl3Dgaussians}.
We find that the volume maximization term in \refEq{our_training_objective} only needs to be applied sporadically.
In practice, we set $\lambda=1$ every ten training iterations and $\lambda=0$ for all other iterations.

%---------------------------------------------------------------

\subsection{Differentiable Uncertainty Optimization}
\label{sec:differentiable_uncertainty_optimization}

To obtain uncertainty estimates per view, we incorporate \refEq{manifold_model} into \refEq{mc_uncertainty}, yielding
\begin{equation}
\label{eq:our_uncertainty}
    \uncertainty_\textrm{manif}(\image(\auxcoords))
    =
    \sum_{\ray \in \image}
    \frac{1}{\nummcsamples}
    \sum_{\manifoldvolumesamples \sim \mathcal{U}( \left[-1, 1 \right]^\manifolddim )}
    \left\|
    \pixelcolor_{\generator(\manifoldvolumesamples)}(\ray, \auxcoords)
    -
    \bar{\pixelcolor}(\ray, \auxcoords)
    \right\|^2,
\end{equation}
with
\begin{equation}
    \bar{\pixelcolor}(\ray, \auxcoords)
    =
    \frac{1}{\nummcsamples}
    \sum_{\manifoldvolumesamples \sim \mathcal{U}( \left[-1, 1 \right]^\manifolddim )}
    \pixelcolor_{\generator(\manifoldvolumesamples)}(\ray, \auxcoords).
\end{equation}
Similar to the training loss in \refEq{our_training_objective}, we use a \citet{sobol1967distribution} sequence for sampling \manifoldvolumesamples in the unit hypercube, but found $\nummcsamples=2$ necessary for obtaining stable results.
%\TODO{8 for selection}

Importantly, \refEq{our_uncertainty} is trivially differentiable with respect to the camera position (via the rays \ray) and the auxiliary parameters \auxcoords.
This property allows us to conduct gradient-based optimization with respect to these quantities for finding next-best views, for which we use the Adam~\cite{KingBa15} optimizer with default parameters.

%% file: sec/5_results.tex
\section{Experiments}
\label{sec:experiments}

In this section, we evaluate our manifold sampling approach for uncertainty estimation, both quantitatively and qualitatively.
Since ground-truth uncertainty is unknown, we evaluate performance on two distinct tasks:
First, we consider active camera planning (\refSec{camera_planning}), where our objective is to optimize a sequence of cameras to effectively reduce reconstruction uncertainty in ordinary radiance fields.
Second, we explore the task of active illumination planning (\refSec{illumination_planning}). 
Here, our aim is to reconstruct a relightable radiance field, and we seek to optimize a sequence of illumination conditions to be utilized for training.
\rev{We evaluate our uncertainty quantification in \refSec{uncertainty_quantification}.
Finally, we provide an analysis of several further aspects of our approach in \refSec{analysis}.}

%For camera optimization with the synthetic dataset, we want to optimize the next camera position on the upper hemisphere with the highest uncertainty. The camera extrinsic matrix is parameterized by a look-at camera matrix from its center. We initialize the camera center by selecting the Euclidean farthest point from the cameras in the training pool. Then we optimize the camera center for 200 iterations by maximizing the rendering variance. The Monte Carlo sampling number is \unsure{4}. 

%For light optimization, the initialization of the SH coefficient is the Euclidean farthest point in the SH space. We optimize the SH coefficients by maximizing the relighting variance under \unsure{k} test views for \unsure{200} iterations. The Monte Carlo sampling number is \unsure{4}. 

%---------------------------------------------------------------

\subsection{Task 1 -- Active Camera Planning}
\label{sec:camera_planning}

Here, we apply our approach to the task of finding a scene-specific sequence of views to be used for training that optimally reduces reconstruction uncertainty in radiance fields.
Our evaluation protocol is structured as follows:
During the training of a radiance field, we systematically introduce one camera at a time, each added after every 2000 training iterations, starting with a single camera.

We compare different approaches for determining the next-best views.
We begin with a baseline method that employs farthest-point sampling. 
In this approach, a camera is selected from a candidate pool based on its maximum distance from the cameras already in the training set.
Then, we consider the two state-of-the-art uncertainty estimation approaches ActiveNeRF~\cite{pan2022activenerf} and FisherRF~\cite{Jiang2024FisherRF}, as well as the active camera placement (ACP) approach of \citet{kopanas2023}.
We adapted all of the aforementioned approaches to the 3DGS representation. 
Note that none of these approaches support differentiable optimization for the next-best camera; instead, they rely on selecting candidates from a predefined pool.
We do not consider the most recent work on radiance field uncertainty estimation of \citet{goli2023}, as they \rev{use the same Laplace approximation method as FisherRF~\cite{Jiang2024FisherRF} but with different representations: NeRF and 3DGS. To ensure fairness, we reimplemented all baselines with 3DGS and found FisherRF is identical to  Bayes' Rays with 3DGS in computing radiance field parameter uncertainty.
 Further, Bayes' Rays does not model the analytical predictive distribution of rendered images.}
Finally, we consider three variants of our method.
In the first variant, Ours (Sel.), we evaluate \refEq{our_uncertainty} for every candidate view and select the one with the highest uncertainty, restricting our method to match the capabilities of the baselines.
In the second variant, Ours (Opt. Sel.), we use \refEq{our_uncertainty} to differentiably optimize for the next-best view, initialized from the solution of the first variant. 
In our final variant, Ours (Opt. Rnd.), we explore the capabilities of our optimization when the process is initialized from a random view.

We conduct our evaluation on the NeRF Synthetic~\cite{mildenhall2020nerf} and the Mip-NeRF360~\cite{barron2022mip} datasets.
In both cases, the candidate pool consists of the full set of training images available (100 and 150, respectively). 
Since we cannot rely on SfM points, we initialize the 3DGS model using a random point cloud.
\rev{We evaluate the variants Ours (Opt. Sel.) and Ours (Opt. Rnd.) only on the synthetic datasets, as these strategies require capturing new training images online after each next-best view optimization. 
For the Mip-NeRF360 dataset, we are limited to the existing training views and hence only evaluate the variant Ours (Sel.).}
In \refTab{cam_selection_synthetic} and \refTab{cam_selection_real}, we provide the results of a quantitative evaluation of reconstruction accuracy on a held-out set of test views. 
\rev{In \refTab{cam_selection_synthetic}, we report the mean and standard deviation over ten runs, averaged across eight scenes, to demonstrate the stability of our method and the chosen sample count.}
\refFig{cam_selection_synthetic} and \refFig{cam_selection_real} show corresponding qualitative results.
Our approach outperforms the baselines across all metrics for both datasets. Candidate selection based on our uncertainty (Ours Sel.) already surpasses the state-of-the-art view synthesis quality. 
Further, using this selection as a starting point for fine-grained optimization (Ours Opt. Sel.) tends to enhance the quality even more. 
Starting with a random initialization (Ours Opt. Rnd.) yields nearly equal-quality results, indicating that our optimization method avoids getting stuck in local minima.

\rev{In \refTab{upper_bound}, we compare the upper bound of novel-view synthesis quality between 3DGS~\cite{kerbl3Dgaussians} and our method on the NeRF Synthetic dataset, using all training views over 7000 iterations. 
Our results, achieved through a stochastic training scheme, demonstrate comparable quality to the original 3DGS method.}

\begin{table*}
	\begin{center}	
    	\caption
    	{
    		Numerical evaluation for novel-view synthesis with \textbf{active camera selection} on the \textbf{NeRF Synthetic dataset}~\cite{mildenhall2020nerf}. 
    	}
        \vspace{-2mm}
        \label{tab:cam_selection_synthetic}
		\begin{tabular}{lrrrrrrr}
            \toprule
             \multirow{2}{*}[-0.5ex]{Method}
			 & \multicolumn{3}{c}{5 cameras} & \multicolumn{3}{c}{10 cameras}  \\
            \cmidrule(lr){2-4}   \cmidrule(lr){5-7}
			 & PSNR$\uparrow$ & SSIM$\uparrow$ &  LPIPS$\downarrow$  & PSNR$\uparrow$ & SSIM$\uparrow$ &  LPIPS$\downarrow$   \\
			\midrule
			%
			% \multirow{6}{*}[-0.5ex]{\makecell{3D Gaussians \\ \cite{kerbl3Dgaussians}}}  
            %\rowcolor{aliceblue}
			%
            Farthest Point  &21.91  $\pm$ 0.03  &0.836  $\pm$ 0.001 &0.139  $\pm$ 0.001  &25.77  $\pm$ 0.05 &0.899  $\pm$ 0.001 &0.087 $\pm$ 0.001 \\
            %\rowcolor{aliceblue}
			%            
             ActiveNeRF  &21.87  $\pm$ 0.48 &0.831 $\pm$ 0.010 &0.144 $\pm$ 0.010 &25.59 $\pm$ 0.73 &0.891 $\pm$ 0.011  &0.093 $\pm$ 0.009\\
             %\rowcolor{aliceblue}
			%
			 FisherRF  &21.24 $\pm$ 0.60  &0.821 $\pm$ 0.013 &0.151 $\pm$ 0.012 &25.89 $\pm$ 0.69  &0.899  $\pm$ 0.008 &0.087  $\pm$ 0.007 \\
            %\rowcolor{aliceblue}
			%    
			 ACP &21.09 $\pm$ 0.39 &0.818 $\pm$ 0.010 &0.156 $\pm$ 0.009 &26.20 $\pm$ 0.36  &0.902 $\pm$ 0.005  &0.084 $\pm$ 0.004 \\
            %\rowcolor{aliceblue}
			%
			 \textbf{Ours (Sel.)}   & 22.38 $\pm$  0.36 & \textbf{0.841 $\pm$} 0.007 & \textbf{0.136}  $\pm$ 0.006 & 26.99 $\pm$  0.37  & 0.911 $\pm$ 0.003   & 0.077 $\pm$ 0.003  \\
            %\rowcolor{aliceblue}
		%
            \textbf{Ours (Opt. Sel.)}  &\textbf{22.45}  $\pm$  0.50 &\textbf{0.841}  $\pm$ 0.009  &0.140  $\pm$  0.008 & \textbf{27.52}  $\pm$  0.45  & \textbf{0.917}   $\pm$  0.005  & \textbf{0.076}   $\pm$  0.004 \\
            %\rowcolor{aliceblue}
		%
            \textbf{Ours (Opt. Rnd.)}  &22.02  $\pm$ 0.70  &0.839  $\pm$ 0.015 &0.147  $\pm$ 0.012 & 26.78  $\pm$ 0.52  & 0.908  $\pm$ 0.011   & 0.082  $\pm$ 0.008 \\
            %
            %
			%
   %          \multirow{6}{*}[-0.5ex]{\makecell{Plenoxels \\ \cite{fridovich2022plenoxels}}}  
   %          & Farthest Point & & & & & & \\
   %          & ActiveNeRF \cite{pan2022activenerf} & & & & & & \\
			% & FisherRF \cite{Jiang2024FisherRF} & & & & & & \\
			% & \cite{kopanas2023} & & & & & & \\
			% & \textbf{Ours -- Candidate Selection} & \textbf{} & \textbf{}   & \textbf{}  & \textbf{} & \textbf{}   & \textbf{}  \\
   %          & \textbf{Ours -- Optimization} & \textbf{} & \textbf{}   & \textbf{}  & \textbf{} & \textbf{}   & \textbf{}  \\
			%
			\bottomrule
		\end{tabular}
	\end{center}
\end{table*}

\begin{table}[!t]
    \setlength{\tabcolsep}{3pt}
	\begin{center}
    	\caption
    	{
            Numerical evaluation for novel-view synthesis with \textbf{active camera selection} on the \textbf{Mip-NeRF360 dataset}~\cite{barron2022mip}.
    	}
        \vspace{-2mm}
        \label{tab:cam_selection_real}	
		\begin{tabular}{lrrrrrr}
            \toprule
            \multirow{2}{*}[-0.5ex]{Method}
			  & \multicolumn{3}{c}{10 cameras} & \multicolumn{3}{c}{20 cameras} \\
            \cmidrule(lr){2-4} \cmidrule(lr){5-7}
			  & PSNR$\uparrow$ & SSIM$\uparrow$ &  LPIPS$\downarrow$  & PSNR$\uparrow$ & SSIM$\uparrow$ & LPIPS$\downarrow$  \\
			\midrule
			ActiveNeRF  &12.50 &0.265 &0.615 &14.38 &0.368 &0.569 \\
			FisherRF &18.02 &0.550 &0.408 &21.38 &0.673 &0.330 \\
			ACP &18.71 &0.568 &0.399 &21.41 &0.684 &0.320 \\
			\textbf{Ours (Sel.)} &\textbf{18.87} &\textbf{0.578} &\textbf{0.388} &\textbf{22.27} &\textbf{0.696} &\textbf{0.312}  \\
			\bottomrule
		\end{tabular}
	\end{center}
\end{table}

\begin{table}[!t]
	\begin{center}
    	\caption
    	{
            \rev{Numerical evaluation for novel-view synthesis with \textbf{all training cameras} on the \textbf{NeRF Synthetic dataset}~\cite{mildenhall2020nerf}.}
    	}
        \vspace{-2mm}
        \label{tab:upper_bound}	
		\begin{tabular}{lrrr}
            \toprule
              
	        Method & PSNR$\uparrow$ & SSIM$\uparrow$ &  LPIPS$\downarrow$   \\
			\midrule
                %\rowcolor{aliceblue}
			%
			3DGS &32.58   &0.966    &0.041   \\
                %\rowcolor{aliceblue}
			%
			\textbf{Ours} &32.11    &0.963   &0.045 \\

			\bottomrule
		\end{tabular}
	\end{center}
\end{table}

%---------------------------------------------------------------

\subsection{Task 2 -- Active Illumination Planning}
\label{sec:illumination_planning}

As a second application domain of our approach, we choose the task of training a relightable radiance field~\cite{martin2021nerf,nerv2021} using controlled illumination.
Here, we focus on a scenario where scenes are captured from multiple viewpoints under various illumination conditions, which can be achieved, for instance, through a light stage setup~\cite{debevec2000acquiring}.
Rather than optimizing for emitted radiance using a set of $m$ spherical harmonics (SH) coefficients per primitive, as commonly done, we take inspiration from precomputed radiance transfer techniques~\cite{sloan_prt,lyu2022nrtf,saito2024rgca}. 
In our method, we optimize for a radiance transfer matrix $\transfermatrix \in \mathds{R}^{m \times n}$ for each primitive in the scene. 
Distant illumination is parameterized by an SH coefficient vector $\auxcoords \in \mathds{R}^n$. 
This vector is multiplied with each transfer matrix \transfermatrix to compute the outgoing radiance per Gaussian that needs to be rendered. 
This setup allows us to efficiently represent and manipulate illumination conditions for relightable rendering.
In our experiments, we set $m=n=16$, covering the first four SH bands.

Analogous to the setup in \refSec{camera_planning}, we start with a single illumination condition -- we choose an angularly uniform illumination, \ie only the DC coefficient is active -- and gradually add new illumination samples every 7000 training iterations.
The samples are obtained by differentiably optimizing uncertainty per \refEq{our_uncertainty} with respect to the illumination condition \auxcoords.
The optimization is initialized from the one-hot parameter vector \auxcoords that results in the highest uncertainty.
Note how our gradient-based approach avoids excessive enumeration of candidates in the 16-dimensional parameter space.

In \refFig{results_light_syn} and \refFig{relighting_quant} we compare our approach against random and \citet{sobol1967distribution} sampling of next illumination conditions using three scenes from the NeRF Synthetic dataset.
We see that our uncertainty-guided approach significantly outperforms the scene-agnostic baselines.

%---------------------------------------------------------------

\subsection{Uncertainty Quantification}
\label{sec:uncertainty_quantification}

\rev{We evaluate pixel-wise predicted uncertainty in the sparse-view setting by analyzing its correlation with depth estimation error, which serves as an indicator of underlying geometric uncertainty.
Specifically, we follow \citet{shen2022conditional} and use the Area Under Sparsification Error (AUSE) metric on the LF dataset~\cite{yucer2016efficient}. 
The results of this analysis are presented in \refTab{AUSE}, where we compare our approach against CFNeRF~\cite{shen2022conditional}, Bayes’ Rays~\cite{goli2023}, and FisherRF~\cite{Jiang2024FisherRF}. 
The results for CFNeRF and Bayes’ Rays are taken directly from \citet{goli2023}.}

%\rev{We  evaluate our method with the Area Under Sparsification Error (AUSE) metric on the LF~\cite{yucer2016efficient} dataset in \refTab{AUSE}, comparing against CFNeRF~\cite{shen2022conditional}, Bayes’ Rays~\cite{goli2023}, and FisherRF~\cite{Jiang2024FisherRF}. 
%Each test image has its pixels removed twice: first, based on the absolute error with ground truth depth, and second, according to an uncertainty measure. The Mean Absolute depth Error of the remaining pixels is calculated for both processes, producing two error curves. The area (difference) between these two curves is the Area Under the Sparsification Error (AUSE), which quantifies the correlation between the predicted uncertainties and the actual errors in the depth predictions. The dataset split is the same as the original  CFNeRF.  Results for CFNeRF and Bayes’ Rays are taken directly from~\citet{goli2023}.}

\rev{Our method can only be fairly compared against FisherRF, as it is the only baseline using the 3DGS representation. 
The potential superiority of Bayes’ Rays in AUSE may be due to limitations of 3DGS with forward-facing scenes and NeRF’s inherent ability to ensure smoothness in the 3D uncertainty field, which leads to smoother depth errors. 
To enable a fairer comparison, future work will extend our method to NeRF-based architectures. 
Our stochastic radiance field and training methodology are also readily applicable to discrete volume representations, such as Plenoxels~\cite{fridovich2022plenoxels}. 
Additionally, future work will explore extending manifold sampling to general continuous neural representations.}

\begin{table}[!t]
	\begin{center}
    	\caption
    	{
    		\rev{Area Under Sparsification Error (AUSE$\downarrow$) on the LF dataset~\cite{yucer2016efficient}.} 
    	}
        \vspace{-2mm}
        \label{tab:AUSE}	
		\begin{tabular}{lrrrrr}
            \toprule

			  Method &Africa     & Basket     & Statue       &Torch    &Avg.  \\
			\midrule
			%
                %\rowcolor{aliceblue}
                CFNeRF w/ NeRF       &0.35               &0.31      &0.46        & 0.97    &0.52 \\
                %\rowcolor{aliceblue}
                Bayes’ Rays w/ NeRF    &\textbf{0.27}          &\textbf{0.28}       & \textbf{0.17}        &\textbf{0.22}   &0.23 \\
                %\rowcolor{aliceblue}
                FisherRF w/ 3DGS      &0.64           & 0.54        & 0.52       & 0.47    & 0.54 \\
                %\rowcolor{aliceblue}
                Ours w/ 3DGS            &\textbf{0.27}                 &0.44       & 0.47        &0.44    & 0.40 \\

			\bottomrule
		\end{tabular}
	\end{center}
\end{table}

%---------------------------------------------------------------

\subsection{Analysis}
\label{sec:analysis}

Here, we analyze various aspects of our approach.
First, we examine different methods for modeling parameter covariance as shown in \refTab{ablation}. 
Since a na\"ive parameterization of the full $\paramdim \times \paramdim$ covariance matrix is generally intractable (it would require petabytes of memory for a typical scene), we explore the tractable alternatives listed in \refFig{sampling_strategies}, including our approach with different ranks \manifolddim.
We consider the task of active camera planning (\refSec{camera_planning}) and measure novel-view synthesis quality on the NeRF Synthetic dataset.
Additionally, we report the number of milliseconds required for one training iteration; the total number of training iterations required to reach a converged radiance field is roughly the same across methods.

We see that low-rank approximations consistently outperform the alternatives.
Surprisingly, the rank has only a minor impact on result quality, but higher ranks increase training time. Our choice of $\manifolddim=2$ strikes a reasonable balance between quality and speed.
Specifically, our solution is only 14\% slower to train than a vanilla 3DGS model without uncertainty (w/o \uncertainty), on top of which a Laplace approximation might be run. 
\rev{Asymptotic time complexity can be computed by analyzing the sampling in \refEq{manifold_model} and the rasterization of Gaussians. \refEq{manifold_model} leads to complexity 
$\mathcal{O}(\nummcsamples N \manifolddim)$ 
while rasterization is 
$\mathcal{O}(\nummcsamples N (p+ \log N))$, 
where \nummcsamples is the number of samples, $N$ is the number of Gaussians, \manifolddim is the rank, and $p$ is the number of pixels.}

In \refFig{energies}, we show representative examples of uncertainty landscapes generated by different methods when a single training view (red dot) is available. 
FisherRF provides a smooth landscape, but optimizing camera positions with this method is impractical due to the need for higher-order derivatives. 
ACP's energy landscape is fairly uniform, as it aims to prevent view clustering, which is not a particularly useful measure in a sparse-view setting.
Again, this energy cannot be differentiated w.r.t. camera position.
Stochastic sampling with a diagonal or block-diagonal covariance matrix introduces excessive noise, whereas our low-rank solutions are smooth by construction.
To gain further insights into the optimization dynamics, we start from a random initialization (yellow dot) and differentiably optimize for a view that maximizes uncertainty (green dot).
With diagonal and block-diagonal covariance matrices, this optimization eventually converges, but it requires many iterations due the noisy uncertainty landscape. 
In contrast, our manifold sampling stably converges two to three times faster, even in cases where the total trajectory is longer.

\begin{table}[!t]
    \setlength{\tabcolsep}{3pt}
	\begin{center}
    	\caption
    	{
    		Ablation.
            We consider different ways to model covariance.
            In addition to novel-view synthesis quality based on 5 and 10 training views, we report the number of milliseconds required for one training iteration.
    	}
        \vspace{-2mm}
        \label{tab:ablation}	
		\begin{tabular}{lrrrrrrr}
            \toprule
            \multirow{2}{*}{Method} & \multirow{2}{*}{Time}
            % Time(s/1000 iter.)
			  & \multicolumn{3}{c}{5 cameras} & \multicolumn{3}{c}{10 cameras} \\
            \cmidrule(lr){3-5} \cmidrule(lr){6-8}
			  & & PSNR$\uparrow$ & SSIM$\uparrow$ &  LPIPS$\downarrow$  & PSNR$\uparrow$ & SSIM$\uparrow$ & LPIPS$\downarrow$  \\
			\midrule
            w/o \uncertainty &  10.8 & --- & --- & --- & --- & --- & --- \\
            \midrule
            Diag. &14.0 &21.54 &0.823 &0.151 &26.91 &0.911 &0.077 \\
            Block-D. &13.8 &21.85  &0.830   &0.145   &26.09   &0.900    &0.082 \\
      
			Rank 1 & 11.6 &22.02 &0.836 &0.139 & \textbf{27.09} &\textbf{0.913}&\textbf{0.076}  \\
			Rank 2 &12.3 & \textbf{22.19} & 0.840 & 0.135 & 26.69 & 0.909   & 0.078  \\
            Rank 4 & 13.5 & 22.17 & \textbf{0.842} & \textbf{0.133} &  26.68  & 0.907   & 0.079 \\
            Rank 10 &15.7 &22.12 &0.834   &0.142   &26.08   & 0.899   &0.086    \\
			\bottomrule
		\end{tabular}
	\end{center}
\end{table}

%% file: sec/6_conclusion.tex
\section{Conclusion}
We proposed a method that explicitly accounts for the epistemic uncertainty in Gaussian radiance fields.
We observed that uncertainty can be modeled as a low dimensional manifold in parameter space, which allows efficient Monte Carlo sampling.
Importantly, our formulation is fully differentiable, which in contrast to most prior works, allows continuously optimizing scene parameters such as the next-best camera view. 
We demonstrated the versatility of our formulation through various experiments, with a focus on active camera and illumination planning.
%
%\rev{Due to the sparsity of the 3DGS representation, our method struggles to model smooth pixel-wise uncertainty, as illustrated in  \refFig{teaser} and \refSec{uncertainty_quantification}.} 
In the future, we plan to investigate the applicability of manifold sampling to other radiance field representations and even to different signal modalities.

\begin{figure*}
\includegraphics[width=\linewidth]{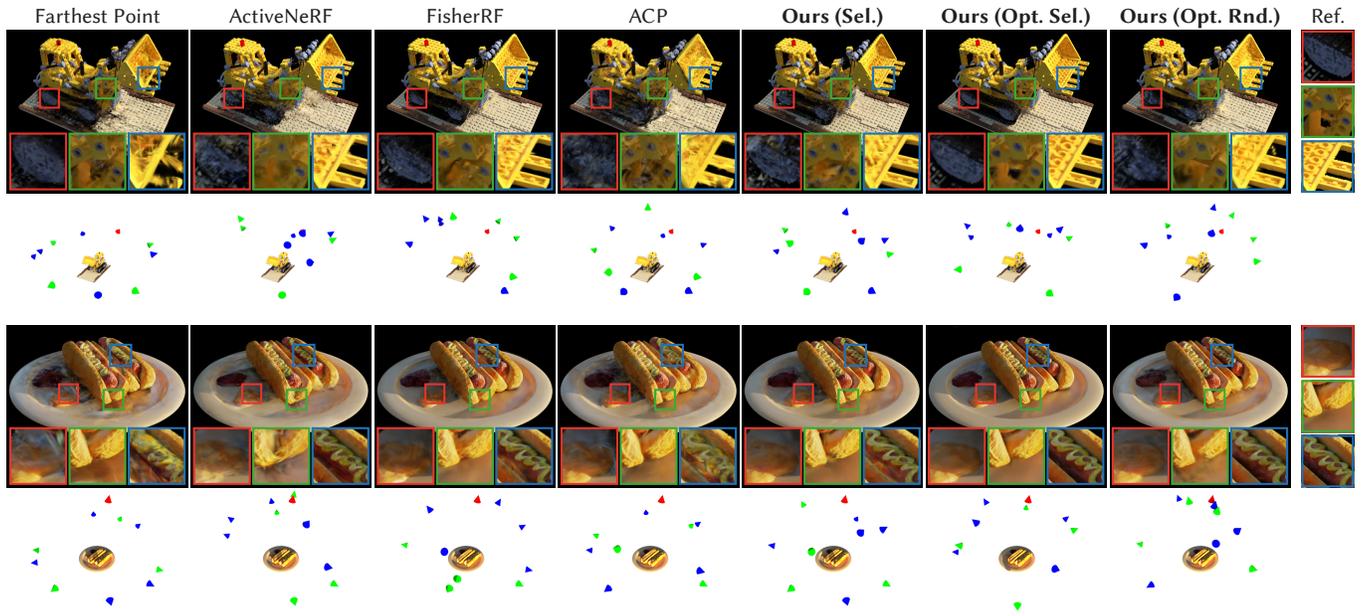} 
\caption
{
Active camera planning results on the NeRF Synthetic dataset from different methods (columns) using 10 training views.
The first row of each scene shows a novel view, while the second row illustrates the distribution of cameras placed by each method.
Red cones represent the initial camera position, green cones represent the first four camera positions, and blue cones represent the final five camera positions.
}
\label{fig:cam_selection_synthetic}
\end{figure*}

\begin{figure*}
\includegraphics[width=\linewidth]{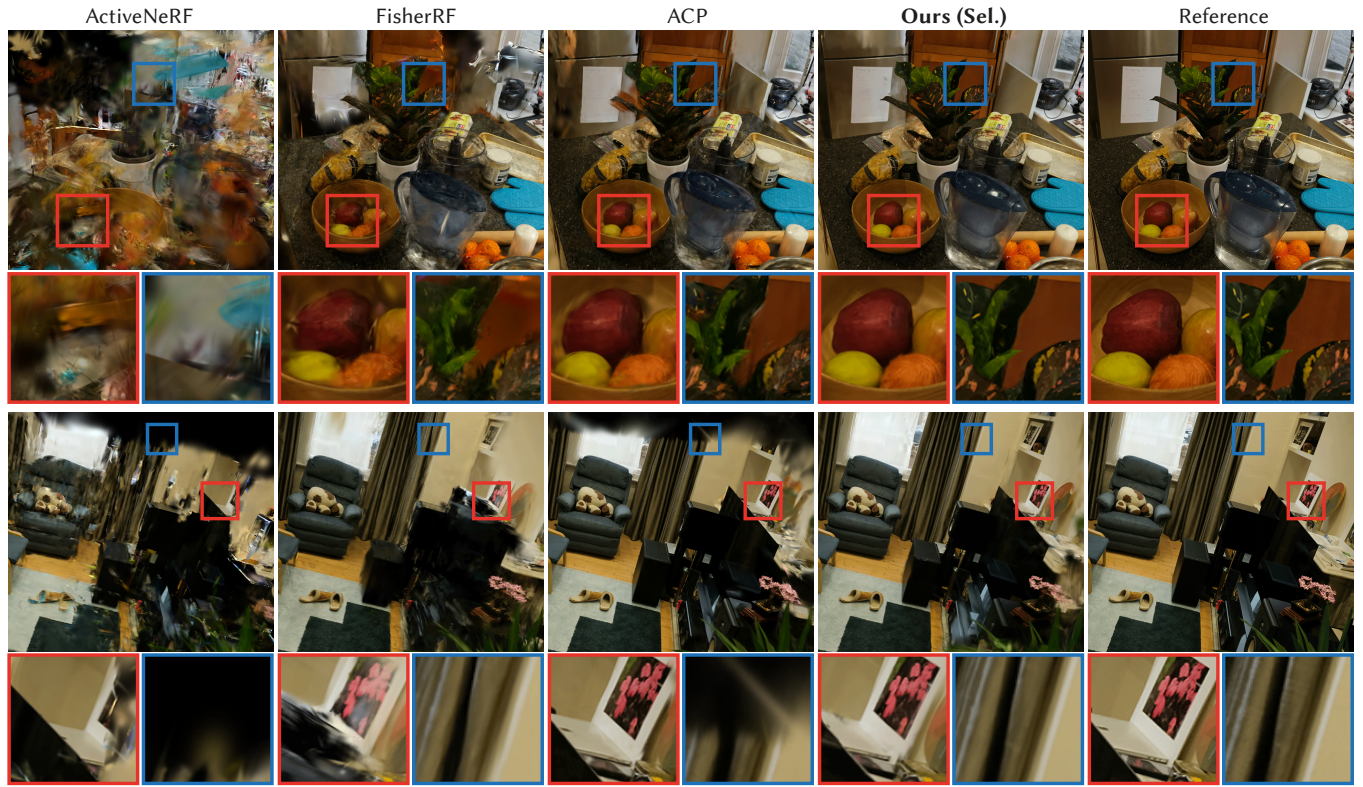} 
\caption
{
Active camera planning results on the MipNeRF-360 dataset from different methods (columns) using 20 training views.
The images show a novel view.
}
\label{fig:cam_selection_real}
\end{figure*}

\begin{figure*}
\includegraphics[width=\linewidth]{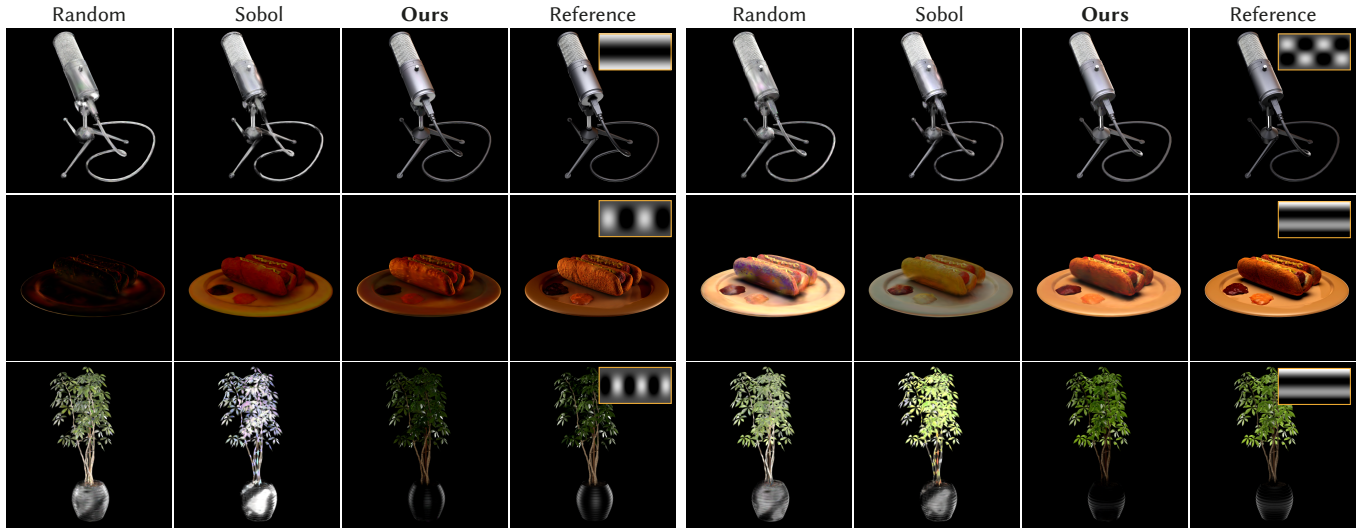} 
\caption
{
Active illumination planning results.
We demonstrate novel-view relighting results based on 8 training illumination conditions.
We display two test illuminations per scene, shown as insets in the respective reference image.
}
\label{fig:results_light_syn}
\end{figure*}

\begin{figure}
\includegraphics[width=\linewidth]{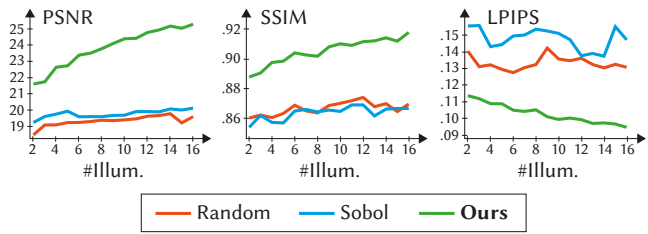} 
\caption
{
Quantitative evaluation of relighting quality.
We plot image quality metrics against the number of illumination conditions used for training.
}
\label{fig:relighting_quant}
\end{figure}

\begin{figure}
\includegraphics[width=\linewidth]{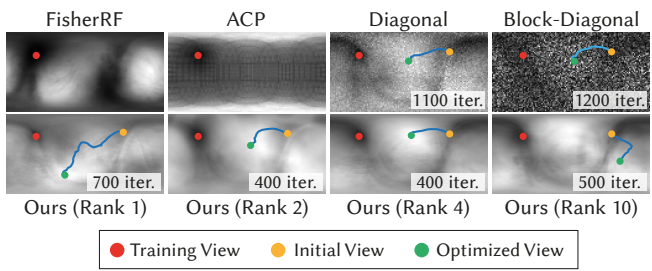} 
\caption
{
Uncertainty landscapes for different methods.
We display uncertainties associated with cameras positioned on a scene-enclosing sphere using a latitude-longitude parameterization.
The only camera in the training pool (red dot) is surrounded by low uncertainty.
Where applicable, we optimize for the second view starting from an initialization (yellow dot) to arrive at a view that maximizes the rendered uncertainty (green dot).
We list the number of optimization iterations required for each approach.
}
\label{fig:energies}
\end{figure}